\documentclass[10 pt, conference]{ieeeconf}  

\IEEEoverridecommandlockouts                

\let\labelindent\relax
\usepackage[shortlabels]{enumitem}
\usepackage{bbm}
\usepackage{times}
\usepackage[utf8]{inputenc} 
\usepackage[T1]{fontenc}    
\usepackage{url}            
\usepackage{booktabs}       
\usepackage{nicefrac}       
\usepackage{microtype}      
\usepackage{adjustbox}      
\usepackage{graphics,color}

\usepackage{dirtytalk}
\makeatletter
\let\NAT@parse\undefined
\makeatother
\usepackage[font=small,labelfont=bf]{caption}
\usepackage{bbm}
\usepackage{bm}
\usepackage{algorithm}
\usepackage[numbers]{natbib}

\usepackage{amsfonts}       
\usepackage{amsmath}       
\usepackage{amssymb}
\usepackage{xspace}

\makeatletter
\DeclareRobustCommand\onedot{\futurelet\@let@token\@onedot}
\def\@onedot{\ifx\@let@token.\else.\null\fi\xspace}
\makeatother

\makeatletter
\newcommand*{\addFileDependency}[1]{
  \typeout{(#1)}
  \@addtofilelist{#1}
  \IfFileExists{#1}{}{\typeout{No file #1.}}
}
\makeatother

\usepackage{xcolor}
\definecolor{ourblue}{rgb}{0.368,0.507,0.71}    
\definecolor{ourorange}{rgb}{0.881,0.611,0.142} 
\definecolor{ourgreen}{rgb}{0.56,0.692,0.195}   
\definecolor{ourred}{rgb}{0.923,0.386,0.209}    
\definecolor{ourviolet}{rgb}{0.528,0.471,0.701} 
\definecolor{ourbrown}{rgb}{0.772,0.432,0.102}  
\definecolor{ourazure}{rgb}{0.364,0.619,0.782}  
\definecolor{ourolive}{rgb}{0.572,0.586,0.}     
\definecolor{ourgray}{RGB}{102,88,84}           

\definecolor{ourblue2}{RGB}{9,134,223} 
\definecolor{ourdarkblue2}{RGB}{5,97,164} 
\definecolor{ourlightblue2}{RGB}{132,201,250} 

\definecolor{ourorange2}{RGB}{224,90,18} 
\definecolor{ourdarkorange2}{RGB}{160,63,9} 
\definecolor{ourlightorange2}{RGB}{246,175,137} 

\definecolor{ouryellow2}{RGB}{227,213,25} 
\definecolor{ourdarkyellow2}{RGB}{177,166,17} 
\definecolor{ourlightyellow2}{RGB}{242,235,140} 

\definecolor{ourpink2}{RGB}{247,24,139} 
\definecolor{ourdarkpink2}{RGB}{164,4,86} 
\definecolor{ourlightpink2}{RGB}{250,163,207} 

\definecolor{ourgreen2}{RGB}{159,198,52} 
\definecolor{ourdarkgreen2}{RGB}{109,138,30} 
\definecolor{ourlightgreen2}{RGB}{209,228,154} 

\definecolor{ourgray2}{RGB}{124,124,115} 
\definecolor{ourdarkgray2}{RGB}{87,87,81} 
\definecolor{ourlightgray2}{RGB}{194,194,189} 

\usepackage{xr-hyper}
\usepackage[pagebackref,breaklinks,colorlinks,allcolors=ourblue]{hyperref}
\usepackage{cleveref}  %

\newcommand{\trsp}{{\scriptscriptstyle\top}}

\DeclareMathOperator*{\argmin}{arg\,min}

\title{\LARGE \bf A Smooth Analytical Formulation of Collision Detection and Rigid Body Dynamics With Contact}

\author{\normalsize
        Onur Beker$^{1}$,
        Nico Gürtler$^{1}$,
        Ji Shi$^{1}$,
        A.\ Ren\'e Geist$^{1}$,
        Amirreza Razmjoo$^{2,3}$,
        Georg Martius$^{1}$,
        Sylvain Calinon$^{2,3}$ \\
        $^{1}$University of Tübingen \quad 
        $^{2}$Idiap Research Institute \quad 
        $^{3}$EPFL
        }

\begin{document}
\twocolumn[{%
\renewcommand\twocolumn[1][]{#1}%
\maketitle

\begin{center}
    \centering
    \captionsetup{type=figure}
    \includegraphics[width=\textwidth]{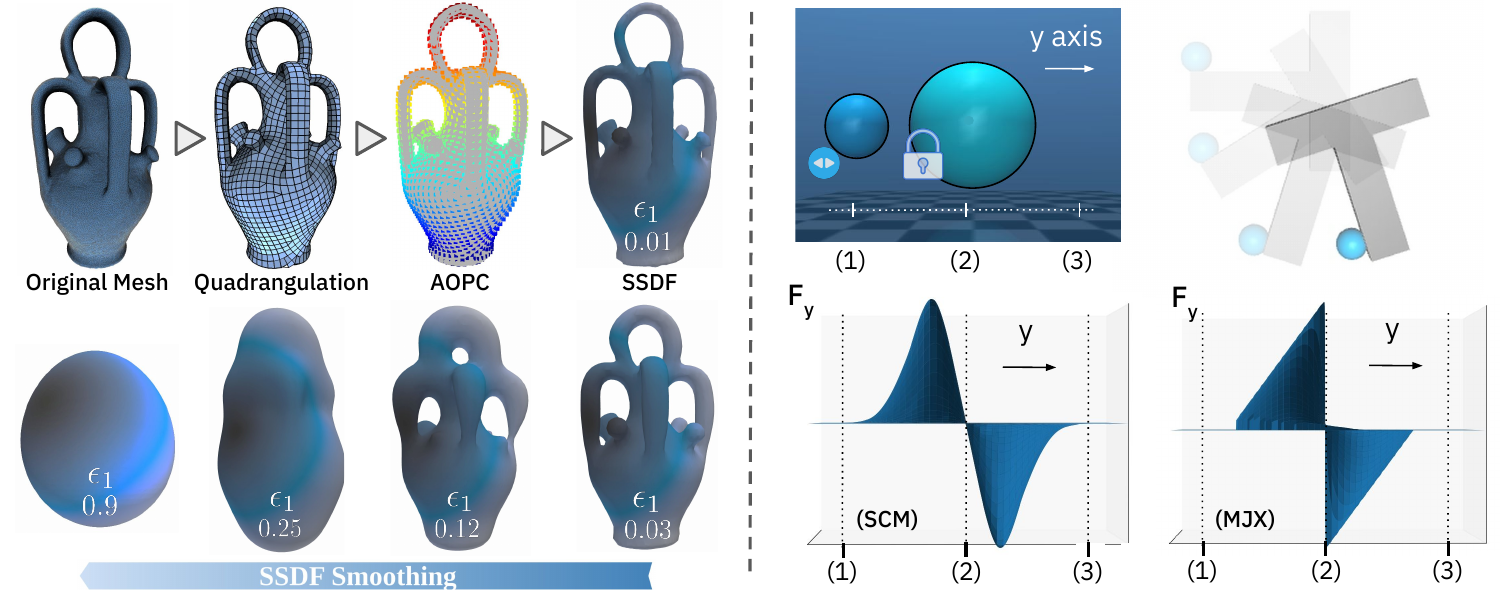}
    \captionof{figure}{We propose a joint formulation for collision detection and contact modelling where both capacities take the form of smooth analytical functions. On the collision detection side (above left), we propose a 3D surface representation called ``soft signed distance function'' (SSDF) that can tightly approximate any closed 3D surface, and continuously interpolate it towards an ellipsoid by means of a smoothing parameter $\epsilon$. SSDF's allow us to formulate a ``soft-minimum contact model'' (SCM) that represents the contact interaction between two collision geometries as a smooth force field distributed over their entire volume of intersection. Right side of the figure visualizes this smoothing effect with an example showing how the contact force between two spheres changes as one is moved continuously along the y axis while the other is fixed, for SCM and contact models from other well-established simulators. We demonstrate the plausibility of the proposed formulation through simulation experiments on a planar pushing system involving a T-shaped collision geometry.}
    \label{fig:title}
\end{center}%
}]

\begin{abstract}
Generating intelligent robot behavior in contact-rich settings is a research problem where zeroth-order methods currently prevail. A major contributor to the success of such methods is their robustness in the face of non-smooth and discontinuous optimization landscapes that are characteristic of contact interactions, yet zeroth-order methods remain computationally inefficient. It is therefore desirable to develop methods for perception, planning and control in contact-rich settings that can achieve further efficiency by making use of first and second order information (i.e., gradients and Hessians). To facilitate this, we present a joint formulation of collision detection and contact modelling which, compared to existing differentiable simulation approaches, provides the following benefits: i) it results in forward and inverse dynamics that are entirely analytical (i.e. do not require solving optimization or root-finding problems with iterative methods) and smooth (i.e. twice differentiable), ii) it supports arbitrary collision geometries without needing a convex decomposition, and iii) its runtime is independent of the number of contacts. Through simulation experiments, we demonstrate the validity of the proposed formulation as a \say{physics for inference} that can facilitate future development of efficient methods to generate intelligent contact-rich behavior.
\end{abstract}

\section{Introduction} \label{intro}
The dynamics of rigid body interactions involving contact is discontinuous in acceleration and velocity due to discrete contact events \cite{anitescu2006optimization, stewart2000rigid, pfeiffer2000multibody, featherstone2014rigid}. 
When modeling such interactions for robotics applications, this inherent non-smoothness gives rise to two distinct objectives: i) prioritizing forward simulation accuracy by capturing the discontinuous reality of contact interactions as faithfully as possible (i.e.\ \say{physics for prediction}), or ii) making appropriate relaxations to this discontinuous reality of contact in order to create a well-behaved optimization landscape (i.e.\ \say{physics for inference}). As these priorities essentially constitute orthogonal complements, 
most well-established simulators \cite{coumans2016pybullet, drake, todorov2012mujoco} try to strike a correct balance between them. 
Following the success of zeroth-order methods in generating intelligent contact-rich behavior (e.g.\ using reinforcement learning \cite{chen2023visual} or stochastic optimal control \cite{howell2022}), development of high-throughput simulators with contact models that are as realistic as possible to facilitate the sim-to-real transfer of such methods has become a particularly relevant research direction \cite{makoviychuk2021isaac, lidec2024end}. Therefore, priorities from the former physics for prediction perspective have in recent years started taking precedence in contact modeling. 
While the zeroth-order approach has certainly proved immensely valuable and effective for the contact-rich manipulation domain, it also remains a fact that most of the efficient and well-understood algorithms from optimal control \cite{borrelli2017predictive}, state estimation \cite{barfoot2024state}, system identification \cite{khalil2004modeling}, and numerical optimization \cite{nocedal1999numerical} that sustain our modern civilization (e.g., making our ships navigate, localizing our satellites, regulating our power grids and manufacturing processes) do operate primarily through using first- and second-order information (i.e., gradients and Hessians). 
This apparent dichotomy raises the following question: is it possible to get the best of both worlds? 
How can we make better use of first- and second-order information to efficiently solve robotic perception, planning, and control problems involving contacts? 
This line of reasoning naturally leads to an earlier question: how can we obtain this first- and second-order information reliably in the context of contact dynamics if it is inherently discontinuous and non-smooth? 
As a step towards answering this latter question and formulating a physics for inference around contact, we introduce the following contributions in this paper:
\begin{itemize}[leftmargin=*, label={\color{ourblue}\textbf{\textbullet}}]
    \item {\color{ourblue}\textbf{Soft signed distance function (SSDF):}} An expression that allows smoothly approximating the signed distance function (SDF) associated with any closed 3D surface (not necessarily convex or simply connected), and continuously interpolating between the original surface and an ellipsoid. We mainly utilize SSDF's to enable analytical and smooth collision detection between surfaces.
    \item {\color{ourblue}\textbf{Soft-minimum contact model (SCM):}} A contact model that uses the SSDF representation to enable forward and inverse dynamics computations in a smooth and analytical manner. This allows the entire simulation pipeline to be written within the confines of the standard feature set (i.e., without any external libraries) of any array computing framework \cite{jax}, easily obtaining gradients and Hessians via automatic differentiation \cite{griewank2008evaluating}, and trivially vectorizing simulations across rigid body systems with \emph{entirely different} geometry as long as they share the same kinematic structure (i.e., rather than copies of the \emph{same} set of geometries in different configurations as in existing simulators \cite{makoviychuk2021isaac}).
\end{itemize}

\section{Related Work}
\subsubsection{Differentiable Physics} Most modern simulators for rigid body dynamics with contact operate through two main steps \cite{anitescu2006optimization, drumwright2011modeling,todorov2014convex, castro_sim}: 
i)~performing collision detection \cite{gilbert1988fast} to find contact points between geometries, and 
ii)~using these contact points to transcribe non-penetration constraints in a convex program that tries to minimize deviation from unconstrained accelerations (i.e., Gauss' principle \cite{redon2002gauss}). 
The resulting dynamics is governed by the KKT system of equations associated with the optimization problem, where contact forces appear as the dual variables for non-penetration constraints. 
A common recipe to differentiate through such simulation pipelines is: 
i)~using a limited set of convex primitives (e.g., ellipsoids, boxes, capsules) with elementary pairwise collision detection routines that are differentiable, and 
ii)~differentiating through the convex program either by explicitly unrolling the solver iterations \cite{franceschi2018bilevel} or by applying the implicit differentiation theorem to its KKT optimality conditions \cite{jaxopt_implicit_diff}. 
Examples of such differentiable simulators include Dojo \cite{howell2022dojo}, the Simple Simulator \cite{lidec2024end},  and the work of \citet{pang2023global}. 
The main differences of our approach are: 
i) our collision detection formulation does not require special convex primitives but can flexibly accommodate any closed 3D shape including non-convex or not simply-connected ones, 
ii) our contact model does not require solving an optimization problem as all steps involved are simple analytical operations. 

\subsubsection{Differentiable Collision Detection} Going beyond elementary geometric arguments between limited shape primitives to more generally support differentiable collision detection between arbitrary surfaces is an active area of research. 
DCOL \cite{tracy2023differentiable} proposes an elegant formulation that supports a highly representative set of convex primitives. 
It operates by solving a cone-constrained convex program to find the minimum scaling factors for each primitive within a collision pair that results in an intersection between them. Gradients through this convex program are then obtained via implicit differentiation. The work by \citet{montaut2023differentiable} in turn proposes a flexible randomized smoothing approach that can support collisions between arbitrary convex sets by: 
i)~sampling $M$ deviations around a nominal kinematic configuration, 
ii)~running GJK and EPA algorithms \cite{gilbert1988fast} $M$ times to get separation distances, 
iii)~getting a zeroth-order estimate of the gradient via the score function estimator \cite{williams1992simple}. 
The difference of our approach with respect to these two approaches is that: 
i)~it also supports non-convex geometries, 
ii)~it does not require any iterative procedure such as solving an optimization problem or running GJK (which itself is a Frank-Wolfe based optimization routine) multiple times. 
\subsubsection{Smoothing Contact Models} A common approach to obtain well-behaved gradients through contact models that involve discontinuities is to find a way to smooth them. The influential work by \cite{pang2023global} proposes a simple and general procedure for smoothing optimization-based contact models (that is interestingly also called \say{analytical} smoothing), which involves moving friction cone constraints into the objective via a log-barrier function and modulating the associated weight (an approach also employed in DCOL \cite{tracy2023differentiable} to obtain smooth derivatives of collision detection). The difference of our formulation is that it proposes an \emph{analytical} smoothing approach instead of modifying an optimization problem that is solved \emph{non-analytically} via an iterative routine.

\section{Methods}
\subsection{Notation}
Let us introduce common notation used in the rest of the paper. We notate generalized positions, velocities, and accelerations \cite{featherstone2014rigid} as $\mathbf{q}, \mathbf{v}, \dot{\mathbf{v}}  \in \mathbb{R}^n$, and use $\bm{p}_i, \bm{v}_i, \bm{a}_i \in \mathbb{R}^3$ to represent translational positions, velocities and accelerations at a point $i$ on a rigid body (all such task space quantities are expressed in world coordinates throughout the rest of the discussion).
The Newton--Euler equation is notated as \cite{featherstone2014rigid}:
\begin{equation}\label{eq:1}
\bm{\tau}  + \sum_i\mathbf{J}_i(\mathbf{q})^\trsp\bm{\lambda}_i = \mathbf{M}(\mathbf{q})\dot{\mathbf{v}} + \mathbf{c}(\mathbf{q}, \mathbf{v})
\end{equation}
\noindent where $\bm{\tau} \in\mathbb{R}^n$ represents the total generalized force due to controls, $\mathbf{M}(\mathbf{q}) \in\mathbb{R}^{n \times n}$ represents the generalized inertia matrix, and $\mathbf{c}(\mathbf{q}, \mathbf{v}) \in\mathbb{R}^n$ represents the total generalized bias force due to effects such as gravity, Coriolis, and centripetal forces. The translational Jacobian at point $i$ is notated as $\mathbf{J}_i(\mathbf{q}) \in \mathbb{R}^{3 \times n}$ and satisfies $\bm{v}_i = \mathbf{J}_i(\mathbf{q})\mathbf{v}$. Finally $\bm{\lambda}_i \in \mathbb{R}^3$ denotes any external linear force acting on point $i$, and $\mathbf{J}_i(\mathbf{q})^\trsp\bm{\lambda}_i$ is the corresponding generalized force induced on the system. Our analytical relaxations to collision detection and contact modelling also make use of the softmax $\bm \sigma$ and softplus $\text{s}^+$ functions, which are defined as:
\begin{align}
    &\bm\sigma_{\epsilon}: \mathbb{R}^N \to [0,1]^N &  \sigma_{\epsilon}(\bm x)_i &= \frac{\exp(x_i / \epsilon)} {\sum_{n=1}^N \exp(x_n/ \epsilon)} \\
    &\text{s}^+_{\epsilon}: \mathbb{R} \to \mathbb{R}_+ & \text{s}^+_{\epsilon}(x) &= \epsilon \log(1 + \exp(x/\epsilon))
\end{align}
where $\epsilon$ acts as a smoothing factor (i.e., as $\epsilon \to 0$, $\bm\sigma_{\epsilon}$ and $\text{s}^+_{\epsilon}$ approach argmax and ReLU functions respectively).

\subsection{Soft Collision Detection}
This section introduces the soft signed distance field (SSDF) representation and describes how to perform collision detection between two closed surfaces represented as SSDFs. 

\begin{figure}[t]
\begin{center}
\includegraphics[width=\columnwidth]{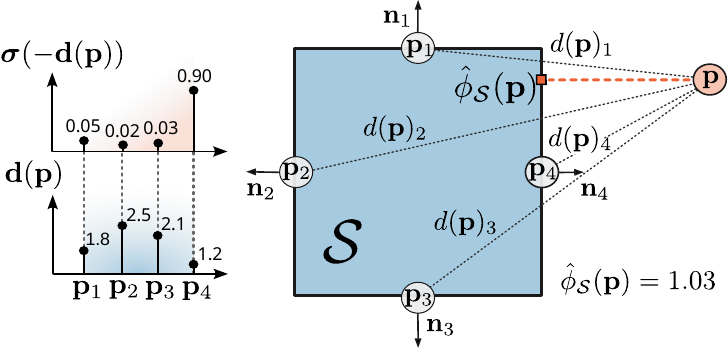}
\end{center}
\caption{Computing the SSDF $\hat{\phi}_\mathcal{S}(\bm{p})$ of a point $\mathbf{p}$ with respect to an AOPC $\mathcal{S}$ involves three steps: i)~computing all distances $\{d(\mathbf{p})_i\}_{\textcolor{ourred}{i=1}}^{\textcolor{ourred}{4}}$ between $\mathbf{p}$ and points $\{\mathbf{p}_i\}_{\textcolor{ourred}{i=1}}^{\textcolor{ourred}{4}}$ on the AOPC, ii)~computing a softminimum $\bm \sigma(-\mathbf{d}(\bm{p}))$ of all distances, resulting in a distribution where the entry $i^*\textcolor{ourred}{=4}$ corresponding to the smallest distance has the largest probability mass, iii)~using the distribution $\bm \sigma(-\mathbf{d}(\bm{p}))$ for soft selection of the corresponding signed distance via a weighted average $\hat{\phi}_\mathcal{S}(\bm{p}) = \sum_{i=1}^I \sigma(-\mathbf{d}(\bm{p}))_i \ \bm{n}_{i}^\trsp(\bm{p} - \bm{p}_{i}) \textcolor{ourred}{\approx \bm{n}_{4}^\trsp(\bm{p} - \bm{p}_{4})}$.}
\label{fig:ssdf_1}
\vspace{-5mm}
\end{figure}

\subsubsection{Representing Geometries} In our framework, geometries are represented with \emph{articulated, oriented point clouds} (AOPCs). An AOPC $\mathcal{S} = \{(\bm{p}_i, \bm{n}_i, \mathbf{J}_i)\}_{i=0}^{I}$ is a collection of 3D planes that cover the surface of a geometry where every plane has a center point $\bm{p}_i$, surface normal $\mathbf{n}_i$, and translational Jacobian $\mathbf{J}_i$ at $\bm{p}_i$. The exact placements of these 3D planes (that are implicitly represented by $\bm{p}_i$ and $\bm{n}_i$) has a significant impact on the numerical conditioning and quality of the analytical approximations in the next sections. A correct placement should distribute them on the surface with high isotropy and in a way that the edges and corners defined by the intersections of these 3D planes align well with the edges and corners of the underlying geometry. In practice, any mesh can be correctly converted to an AOPC by: i) using the remeshing algorithm of \cite{Jakob2015Instant} to quadrangulate the mesh down to a desired resolution, ii) setting $\bm{p}_i$ and $\bm{n}_i$ as the center point and surface normal of each quadrangle, iii) computing $\mathbf{J}_i$ using forward kinematics.

\subsubsection{Soft Signed Distance Function (SSDF)} 
After converting a mesh to an AOPC, we can proceed with making a smooth approximation to its signed distance function. 
Let $\mathbb{S}$ denote the surface of a mesh with normals $n_\mathbb{S}(\bm{x}) \in \mathbb{R}^3$ for any 3D point $\bm{x}$ on $\mathbb{S}$, and $\mathcal{S}$ be an AOPC obtained from $\mathbb{S}$. One way to compute the signed distance $\phi_\mathbb{S}(\bm{p}) \in \mathbb{R}$ of any 3D point $\bm{p}$ with respect to $\mathbb{S}$ is:
\begin{align}
    \bm{x}^* =& \ \argmin_{\bm{x} \in \mathbb{S}} \|\bm{p} - \bm{x}\|^2 \label{eq:softmin_smoothing_1}\\
    \phi_\mathbb{S}(\bm{p}) =& \ \bm{n}(\bm{x}^*)^\trsp(\bm{p} - \bm{x}^*)
\end{align}
Using $\mathcal{S}$, this can be approximated non-smoothly as:
\begin{align}
    i^* =& \ \argmin_{i \in 1, ..., I} \ \|\bm{p} - \bm{p}_i\|^2 \label{eq:sdf_approx_1} \\ 
    \phi_\mathcal{S}(\bm{p}) =& \ \bm{n}_{i^*}^\trsp(\bm{p} - \bm{p}_{i^*}) \label{eq:sdf_approx_2}
\end{align}
and a smooth (i.e. twice differentiable) analytical approximation to (\ref{eq:sdf_approx_1},~\ref{eq:sdf_approx_2}) can in turn be computed as:
\begin{align}
    &d(\bm{p})_i := \|\bm{p} - \bm{p}_i\|^2 \quad\quad ( \text{with} \ \mathbf{d}(\bm{p}) \in \mathbb{R}^I) \\
    &\hat{\phi}_\mathcal{S}(\bm{p}) = \sum_{i=1}^I \sigma_{\epsilon_1}(-\mathbf{d}(\bm{p}))_i \ \bm{n}_{i}^\trsp(\bm{p} - \bm{p}_{i}) \label{eq:softmin_smoothing}
\end{align}

The probability distribution $\bm\sigma_{\epsilon_1}(-\mathbf{d}(\bm{p})) \in [0,1]^I$ effectively evaluates a soft-argminimum of $\|\bm{p} - \bm{p}_i\|$ over all indices $i$, as shown in Fig.~\ref{fig:ssdf_1}. 
Using it to compute a weighted average of $\bm{n}_{i}^\trsp(\bm{p} - \bm{p}_{i})$ in turn approximates $\bm{n}_{i^*}^\trsp(\bm{p} - \bm{p}_{i^*})$. We call $\hat{\phi}_\mathcal{S}(\bm{p})$ the \emph{soft signed distance function} (SSDF). 

For applications that require more fine-grained control on the surface representation, \eqref{eq:softmin_smoothing} can instead be replaced with the following generalization:
\begin{align}
    &\hat{\phi}_\mathcal{S}(\bm{p}) = \frac{\sum_{i=1}^I B_i(\bm{p} - \bm{p}_{i}) \ \bm{n}_{i}^\trsp(\bm{p} - \bm{p}_{i})}{\sum_{i=1}^I B_i(\bm{p} - \bm{p}_{i})}
\end{align}
where $B_i(\bm{p} - \bm{p}_{i})$ are basis functions with local support. For example using an isotropic Gaussian kernel for $B_i$ is equivalent to \eqref{eq:softmin_smoothing}; which in turn is equivalent to implicit moving least squares \cite{kolluri2005provably} as well as the mean of a Gaussian process implicit surface \cite{williams2006gaussian}; and using a non-isotropic Gaussian kernel is equivalent to 3D Gaussian splatting \cite{kerbl20233d}. For our experiments, we have found the isotropic Gaussian kernel with a uniform temperature $\epsilon_1$ across all points to be sufficiently expressive and simple to implement.

\begin{figure}[h]
\begin{center}
\includegraphics[width=0.9\columnwidth]{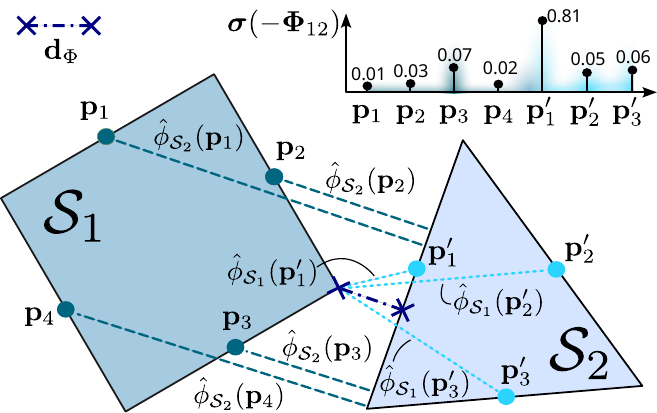}
\end{center}
\caption{Steps involved in performing collision detection between two AOPCs are as follows: i)~computing SSDFs of all points on one AOPC with respect to the other AOPC; and vice versa; which gives $\{\hat{\phi}_{\mathcal{S}_1}(\bm{p}'_{j})\}_{\textcolor{ourred}{j=1:3}}$ and $\{\hat{\phi}_{\mathcal{S}_2}(\bm{p}_{i})\}_{\textcolor{ourred}{i=1:4}}$, ii) stacking them into a single vector $\bm{\Phi}_{{12}} \textcolor{ourred}{\in \mathbb{R}^{7}}$ to form the separation field, iii) computing a soft-argminimum $\bm\sigma(-\bm{\Phi}_{12})\textcolor{ourred}{\in[0, 1]^7}$ over the separation field to obtain a separation distribution, which acts as a soft selection operator whose entries concentrate probability mass on AOPC points that have the largest penetration depth (i.e., the smallest SSDF value).}
\label{fig:ssdf_2}
\vspace{-5mm}
\end{figure}
\subsubsection{Collision Detection Between Two SSDFs}\label{sec:analytica_collision_detection} We now describe how to represent the collision context between two SSDFs to support contact simulation using SCM.
Let $\mathbb{S}_1$ and $\mathbb{S}_2$ be two surfaces, and $\mathcal{S}_1, \mathcal{S}_2$ be the corresponding AOPCs with $I_{1}$ and $I_{2}$ points each. Traditional collision drivers represent the collision context by a separation vector $\bm d_{\Phi} \in \mathbb{R}^3$ that connects the two end-points of the largest penetration distance between two surfaces (i.e., the smallest separation distance in case of non-penetration) \cite{lelidec2023contacts}. These witness points naturally correspond to the pair of points with the minimum signed distance in-between, and therefore the following holds:
\begin{align}\label{eq:sep_distance}
    \| \bm d_{\Phi} \| = \min \ \{\min_{\mathbf{x}'_2 \in \mathbb{S}_2} \phi_{\mathbb{S}_1}(\mathbf{x}'_2) \ ; \ \min_{\mathbf{x}_1 \in \mathbb{S}_1} \phi_{\mathbb{S}_2}(\mathbf{x}_1)\}
\end{align}

Applying a probabilistic relaxation to \eqref{eq:sep_distance} that is analogous to \eqref{eq:softmin_smoothing} , our soft collision detection routine instead characterizes the collision context using an entire separation field distributed over the volume of intersection as follows:
\begin{itemize}[leftmargin=*, label={\color{ourblue}\textbf{\textbullet}}]
    \item Compute $\{\hat{\phi}_{\mathcal{S}_1}(\bm{p}'_j)\}_{j=1}^{I_2}$ and $\{\hat{\phi}_{\mathcal{S}_2}(\bm{p}_i)\}_{i=1}^{I_1}$ (i.e. SSDFs of all points on $\mathcal{S}_1$ with respect to $\mathcal{S}_2$, and vice versa).
    \item Concatenate these two sets of SSDF's into a single vector to create the \emph{separation field} $\bm{\Phi}_{{12}} \in \mathbb{R}^{I_1+I_2}$, and compute the \emph{separation distribution} $\bm\sigma_{\epsilon_2}(-\bm{\Phi}_{12})\in[0, 1]^{I_1+I_2}$.
\end{itemize}

As visualized in Fig.\ref{fig:ssdf_2}, the entries of $\bm\sigma_{\epsilon_2}(-\bm{\Phi}_{12})$ represent how much each point on one surface penetrates into the other surface (i.e., larger the entry, larger the penetration), and a smooth approximation to the separation distance $\| \bm d_{\Phi} \|$ is simply $\bm\sigma_{\epsilon_2}(-\bm{\Phi}_{12})^\trsp \ \bm{\Phi}_{{12}}$, which is a weighted average exactly analogous to \eqref{eq:softmin_smoothing}. 

If we want to obtain a discrete set of collision points to then propagate to any existing contact model \cite{todorov2014convex, lidec2024end}, we can proceed with an additional reduction of the separation field $\bm{\Phi}_{12}$. 
To achieve this, let us first jointly re-index all points across both AOPCs as $\{\bm{p}_f\}_{f=1}^{I_1 + I_2}$. Because these AOPCs were obtained from a quadrangulation, every point $\bm{p}_f$ belongs to a face $f$. Quadrangulations for $\mathbb{S}_1$ and $\mathbb{S}_2$ also produce associated vertices $\{\bm{y}_i\}_{i=1}^{V_1}$ and $\{\bm{y}'_j\}_{j=1}^{V_2}$. 
Let us jointly index these two sets of vertices as $\{\bm{y}_v\}_{v=1}^{V_1 + V_2}$, and use $\mathbbm{1}_{fv}$ to denote if a vertex $v$ belongs to a face $f$ (i.e., $\mathbbm{1}_{fv} = 1$ if it does, $0$ otherwise). 
We can then transfer the probability masses of the separation distribution $\bm\sigma_{\epsilon_2}(-\bm{\Phi}_{12})$ defined over faces $f$ of the quadrangulation onto a distribution of weights $\mathbf{z} \in \mathbb{R}^{V_1 + V_2}$ that is defined over the vertices $\{\bm{q}_v\}$ instead by using:
\begin{align}
    z_v = \sum_{f=1}^{I_1 + I_2} \mathbbm{1}_{fv} \ \sigma_{\epsilon_2}(-\bm{\Phi}_{12})_f 
\end{align}

Given this distribution $\mathbf{z}$ that captures the penetration depth of each vertex $\{\bm{x}_v\}_{v=1}^{V_1 + V_2}$, we can obtain $K$ discrete collision points $\bm{c}_{1:K} \in \mathbb{R}^3$ via soft top-K selection \cite{softK}. For any vector $\mathbf{z} \in \mathbb{R}^V$, the soft top-K operation produces a matrix $\bm\Gamma^K(\mathbf{z}) \in [0, 1]^{K \times V}$ where the entry $\Gamma^K(\mathbf{z})_{kv} \approx 1$ if $z_v$ is the $k$th largest entry of $\mathbf{z}$, and $\Gamma^K(\mathbf{z})_{kv} \approx 0$ otherwise. Using $\bm\Gamma^K(\mathbf{z})$, soft selection then amounts to a weighted average:
\begin{align}
    \bm{c}_{k} \ = \sum_{v=1}^{V_1 + V_2} \Gamma^K(\mathbf{z})_{kv} \ \bm y_v \label{eq:differentiable_contact_points}
\end{align}

When geometries for all collision pairs $(\mathcal{S}_1, \mathcal{S}_2)$ are convex \cite{wei2022coacd} and mesh vertices come from a quadragulation, using $K=4+4 =8$ (e.g., to ensure coverage of non-strictly convex cases like box-box collisions with parallel faces) creates points $\bm{c}_{1:K}$ that are a sufficient description of the collision context analogous to standard collision drivers used in existing simulators \cite{todorov2012mujoco, coumans2016pybullet}. However, reducing the separation field $\bm{\Phi}_{12}$ into $\bm{c}_{1:K}$ may not always be desirable, as $\bm{\Phi}_{12}$ is an even richer description of the collision context that does not require a convex decomposition to be meaningful (e.g., if two non-convex collision geometries establish contact at multiple different points, the distribution $\bm\sigma_{\epsilon_2}(-\bm{\Phi}_{12})$ would simply have multiple modes rather than a single peak). The next section describes a contact model that makes use of the entire separation field.

\subsection{Soft-minimum Contact Model (SCM)}
Building on top of the SSDF representation and the associated soft collision detection routine introduced in the previous sections, this section discusses how they can be used to simulate rigid body dynamics with contact.
\subsubsection{Point-Plane Contact} The main idea of the soft-minimum contact model (SCM) is to compute the total contact force between two SSDFs as a weighted average of the individual contact forces between all possible point-plane pairs across two AOPCs. 
Let us consider a planar face with center $\bm{p}_i$ and normal $\bm{n}_i$, and a moving point at position $\bm{p}$ with velocity $\bm{v}$ decomposed as $\bm{v} = v_n\bm{n}_i + \bm{v}_t$. Following \cite{kurtz2023inverse}, we model the contact force $\bm{\lambda}(\bm{p}, \bm{v}, \bm{p}_i, \bm{n}_i) \in \mathbb{R}^3$ between them using a standard spring-damper system with dissipation velocity $v_d$ and stiction velocity $v_s$:
\begin{align}
    \phi =& \ \bm{n}^\trsp(\bm{p} - \bm{p}_i) \label{eq:point_plane_contact_1} \\
    c(\phi) =& \ k \ s_{\epsilon_3}^+(-\phi) \label{eq:point_plane_contact_2} \\
    d(v_n/v_d) =& \ [v_n/v_d \leq 0] (1-v_n/v_d) \ + \\ & \ [0 < v_n/v_d \leq 2] (v_n/v_d -2)^2/4 \label{eq:point_plane_contact_3} \\
    \lambda_n =& \ c(\phi)d(v_n/v_d) \label{eq:point_plane_contact_4} \\
    \bm{\lambda}_t =& -\mu \lambda_n\frac{\bm{v}_t}{\sqrt{v_s^2 + ||\bm{v}_t||^2}} \label{eq:point_plane_contact_5} \\
    \bm{\lambda}(\bm{p}, \bm{v}, \bm{p}_i, \bm{n}_i) =& \ \lambda_n\bm{n}_i + \bm{\lambda}_t \label{eq:point_plane_contact_6}
\end{align}
\noindent where the Iverson bracket notation $[\text{condition}]$ evaluates to $1$ or $0$ depending on whether the condition holds or not. While we refer the reader to \cite{kurtz2023inverse} for an in-depth explanation, we note that equations \eqref{eq:point_plane_contact_1}--\eqref{eq:point_plane_contact_6} simply construct a spring damper system between a point and the plane where \eqref{eq:point_plane_contact_2} ensures that there is a slight force pushing the point away from the plane even when the two are not in penetration. 
\subsubsection{Point-SSDF Contact} We now describe how to compute contact forces between a point and an SSDF using a weighted average of the point-plane contact forces $\bm{\lambda}(\bm{p}, \bm{v}, \bm{p}_i, \bm{n}_i)$ from the previous section.
Let us consider a moving AOPC $\mathcal{S}$ where every point $\bm p_i$ has an associated linear velocity $\bm{v}_i \in \mathbb{R}^3$. Consider also a moving point (attached to a body) at position $\bm{p}$ with velocity $\bm{v}$ and Jacobian $\mathbf{J}$. The resulting generalized force $\bm{\Lambda}_\mathcal{S}(\bm{p}, \bm{v}, \mathbf{J}) \in \mathbb{R}^n$ (i.e., similar to the $\mathbf{J}_i^\trsp\bm{\lambda}_i \in \mathbb{R}^n$ term in \eqref{eq:1}) can be computed as follows \cite{featherstone2014rigid}:
\begin{align}\label{eq:point_ssdf_force}
&\bm{\bar{\lambda}}_{i} := (\mathbf{J}_i - \mathbf{J})^\trsp\bm{\lambda}(\bm{p}, \bm{v} - \bm{v}_i, \bm{x}_i, \bm{n}_i) \\
&\bm{\Lambda}_\mathcal{S}(\bm{p}, \bm{v}, \mathbf{J}) = \sum_{i=1}^{I} \sigma_{\epsilon_1}(-\mathbf{d}(\bm{p}))_i \ \bm{\bar{\lambda}}_{i} \label{eq:point_ssdf_force_2}
\end{align}

Note that the weighted average in \eqref{eq:point_ssdf_force_2} simply corresponds to a soft selection of the contact force belonging to the point-plane pair with the largest penetration in-between, in a manner directly analogous to \eqref{eq:softmin_smoothing}.
\subsubsection{SSDF-SSDF Contact} Finally, we describe how to compute the total generalized contact force $\bm{\Lambda}_{12}(\mathbf{q}, \mathbf{v})$ between two AOPCs $(\mathcal{S}_1, \mathcal{S}_2)$ as a weighted average of point-SSDF contact forces $\bm{\Lambda}_\mathcal{S}(\bm{p}, \bm{v}, \mathbf{J})$. Analogous to the soft collision detection routine from Sec.~\ref{sec:analytica_collision_detection}, we proceed as follows:
\begin{itemize}[leftmargin=*, label={\color{ourblue}\textbf{\textbullet}}]
    \item Compute contact forces exerted on all points of one SSDF by the other SSDF (and vice versa), creating two lists: $\{\bm{\Lambda}_{\mathcal{S}_1}(\bm{p}'_j, \bm{v}'_j, \mathbf{J}'_j)\}_{j =1}^{I_2}$ and $\{\bm{\Lambda}_{\mathcal{S}_2}(\bm{p}_i, \bm{v}_i, \mathbf{J}_i)\}_{i = 1}^{I_1}$ .
    \item Concatenate these two sets and jointly re-index them as $\{\bm{\Lambda}_{\mathcal{S}_{1 2}}(\bm{p}_f, \bm{v}_f, \mathbf{J}_f)\}_{f=1}^{I_1 + I_2}$.
    \item Compute a weighted average based on the separation distribution computed during collision detection:
    \begin{equation}\label{eq:ssdf_ssdf_force}
        \bm{\Lambda}_{12}(\mathbf{q}, \mathbf{v}) = \sum_{f=1}^{I_1+I_2} \sigma_{\epsilon_2}(-\bm{\Phi}_{12})_f \ \bm{\Lambda}_{\mathcal{S}_{12}}(\bm{p}_f, \bm{v}_f, \mathbf{J}_f)
    \end{equation}
\end{itemize}

The benefit of using an entire separation field $\bm{\Phi}_{12}$ rather than a discrete set of contact points $\mathbf{c}_{1:K}$ to describe the collision context becomes apparent when $(\mathcal{S}_1, \mathcal{S}_2)$ are non-strictly convex (e.g. two boxes lying parallel one on top of the other) or non-convex (e.g. a horseshoe piercing a box with unequal penetration depth on its two ends). Because in such cases \eqref{eq:ssdf_ssdf_force} still remains valid, as the support of the separation distribution simply becomes broader (e.g. the box-box example) or multimodal (e.g., the box-horseshoe example) to cover the entire volume of intersection. This eliminates the need to do convex decompositions or employing contact point buffers to evenly distribute contact forces on a surface. 
\subsubsection{Dynamics Under SCM} Building on top of our definitions from the previous sections about the contact force between two SSDFs, we can now describe how to simulate dynamics using SCM.
Consider a rigid-body system with AOPCs $\{\mathcal{S}_a\}$ and collision pairs $\mathcal{G} = \{(a, b)\}$. The total contact force in configuration space simply becomes a twice-differentiable analytical function $\sum_{(a, b) \in \mathcal{G}} \bm{\Lambda}_{ab}(\mathbf{q}, \mathbf{v})$. Incorporating it into \eqref{eq:1}, forward and inverse dynamics under the soft-minimum contact model simply become:
\begin{align}
    \bm{\tau}(\mathbf{q}, \mathbf{v}, \dot{\mathbf{v}}) =& \ \mathbf{M}(\mathbf{q})\dot{\mathbf{v}} + \mathbf{c}(\mathbf{q}, \mathbf{v}) - \sum_{(a,b) \in \mathcal{G}} \bm{\Lambda}_{ab}(\mathbf{q}, \mathbf{v}) \label{eq:softmin_inv_dyn} \\
    \dot{\mathbf{v}}(\mathbf{q}, \mathbf{v}, \bm{\tau}) =& \ \mathbf{M}(\mathbf{q})^{-1}\Big[\bm{\tau} - \mathbf{c}(\mathbf{q}, \mathbf{v}) + \sum_{(a,b) \in \mathcal{G}} \bm{\Lambda}_{ab}(\mathbf{q}, \mathbf{v})\Big] \label{eq:softmin_fwd_dyn}
\end{align}

\section{Experiments} \label{experiments}
This section presents experiments that demonstrate how SSDFs and rigid body dynamics under SCM behave.
\subsection{Characterizations of the SSDF Isosurfaces}
\begin{figure}[t]
\begin{center}
\includegraphics[width=\columnwidth]{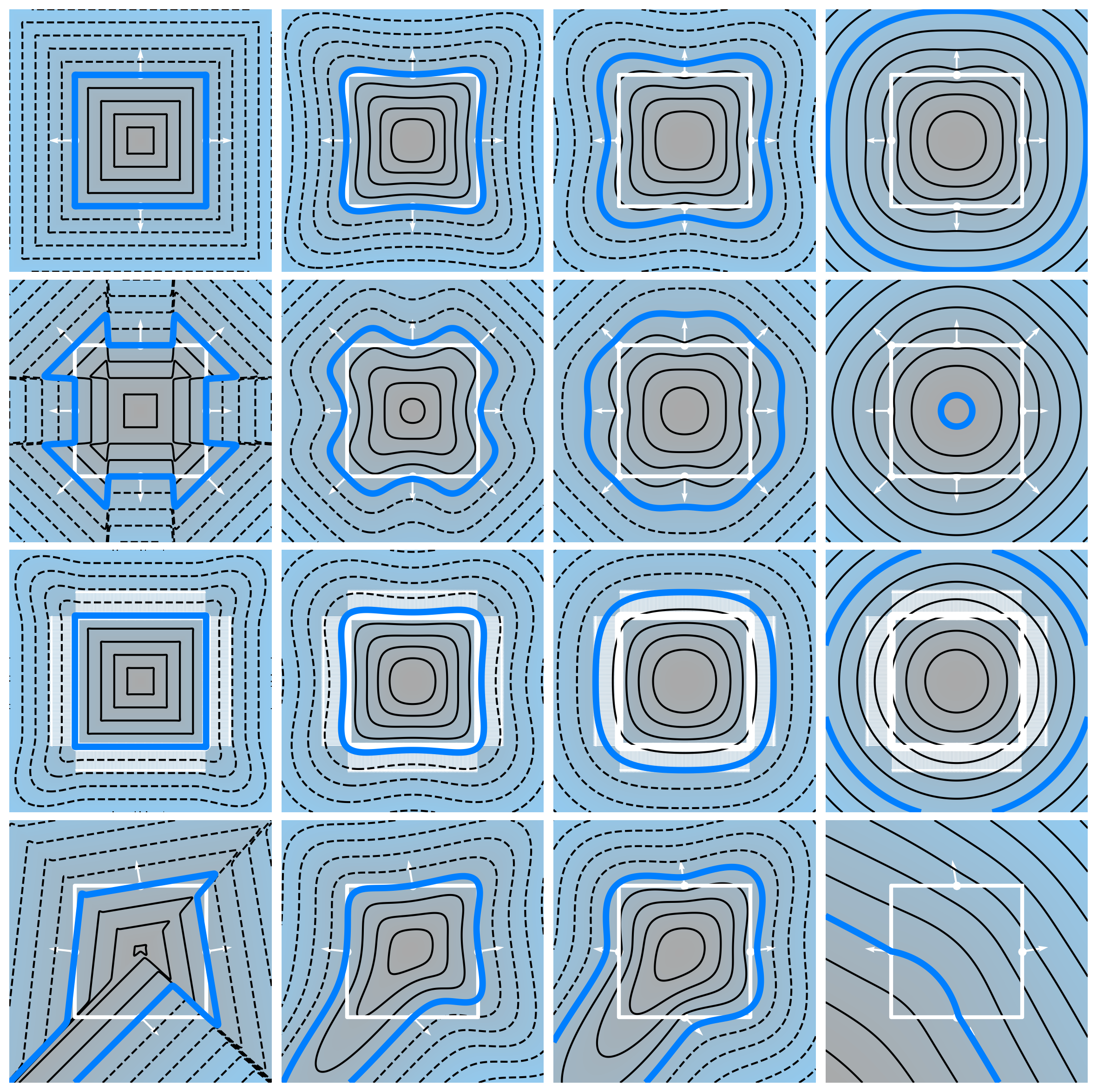}
\footnotesize
\vspace{0.5cm}
$\epsilon = 0.01$ \hspace{1cm} $\epsilon = 0.25$ \hspace{1cm} $\epsilon = 0.5$ \hspace{1cm} $\epsilon = 10.0$\\
\vspace{-5mm}
Softmax temperature
\normalsize
\end{center}
\vspace{-2mm}
\caption{SSDF of a 2D box for different softmax temperatures $\epsilon$, points 
$\bm p_i$, and normals $\bm n_i$. 
The outline of the box, as well as points and normals, are marked in white while the zero isosurface is blue.}
\label{fig:ssdf_box}
\vspace{-5mm}
\end{figure}
We characterize the behavior of the SSDF function using a simple 2D box as a didactic example, as shown in Fig.~\ref{fig:ssdf_box}. First, it can be seen that consistent normal orientation is crucial to obtain a correct SSDF approximation. As the AOPC representation has no explicit face information, it is the normal orientation that defines the boundaries of what constitute the inside and outside of a closed surface, as well as determining how the surface is interpolated between neighboring points. 
Distributing the AOPC points and normals on the surface correctly is also crucial, and it can be seen that additional incorrect points on the corner of the box result in significant artifacts in the SSDF isosurface. 
It can also be seen from Figures \ref{fig:title} and \ref{fig:ssdf_box} that the softmax temperature used in the SSDF approximation essentially acts as a smoothing parameter. As it is decreased from infinity to zero, the isosurface smoothly interpolates between a simply-connected ellipsoid and the original surface, with non-convex regions, holes, and corners gradually appearing. Finally, the resolution of points used in the AOPC is another important hyperparameter, and increasing it suppresses oscillatory artifacts on the SSDF surface that are present when temperature is high (e.g. $\epsilon=0.5$).

\subsection{Characterizations of the Contact Force Profile}
\begin{figure}[h]
\begin{center}
\includegraphics[width=\columnwidth]{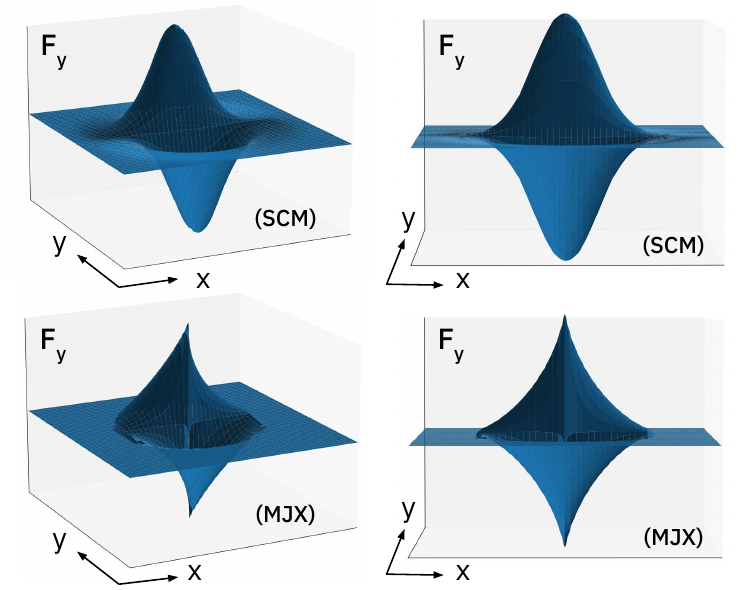}
\vspace{-2.5mm}
\end{center}
\caption{Comparison of contact forces obtained from SCM (top row) and MJX's contact model (bottom row). 
}
\label{fig:scm_force}
\vspace{-2.5mm}
\end{figure}

\begin{figure}[t]
\centering
\includegraphics[width=\columnwidth]{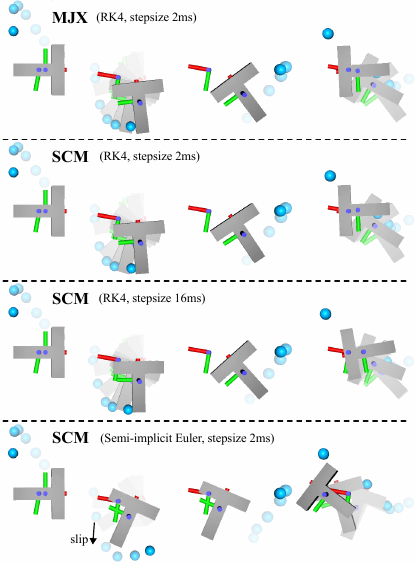}
\caption{Example rollouts for planar pushing experiments. Note that on the last row, the contact at the tip of the object slips rather than sticks due to the inaccuracy in Euler integration. Videos for all rollouts are provided in supplementary material.}
\label{fig:fig_sim}
\vspace{-5mm}
\end{figure}
%
%
%
\begin{figure*}[t]
\centering
\label{tab:results}
\adjustbox{max width=\textwidth}{%
\newcommand{\mc}[1]{\multicolumn{2}{c}{#1}}
\begin{tabular}{@{}l*{8}{r@{$\,\pm\,$}l}@{}}
\toprule
                      & \mc{2-RK-4k}    & \mc{8-RK-4k}   & \mc{16-RK-4k}   & \mc{2-RK-26}   & \mc{8-RK-26}   & \mc{16-RK-26}   & \mc{2-Euler-4k}          & \mc{2-Euler-26}          \\ \midrule
SCM step time (ms)     & 17.9 & 2.6  & 18.6 & 2.6 & 17.7 & 3.2  & 13.5 & 3.1 & 14.6 & 3.2 & 15.6 & 2.8  & 0.9 & 0.3   & 0.8 & 0.1   \\
MJX step time (ms)     & 12.5 & 4.7  & 13.1 & 5.3 & 13.7 & 5.6  & 12.5 & 4.7 & 13.1 & 5.3 & 13.7 & 5.6  & 4.2 & 1.5   & 4.2 & 1.5   \\
$\Delta d$ (cm)       &  1.2 & 1.0   & 1.3 & 1.6 & 4.6  & 3.7  &  2.1 & 0.7 &  2.4 & 1.6  & 6.3 & 5.4  & 6.7 & 2.9   & 11.2& 5.7  \\
$\Delta \theta$ (deg) & 10.8 & 10.3 & 12.5 & 9.1 & 29.4 & 18.3 &  6.8 & 2.7 & 11.4 & 6.5 & 26.6 & 25.3 &42.3 & 18.2 & 46.2 & 34.3 \\ \bottomrule
\end{tabular}%
}
\vspace{2mm}
\caption{Comparison of simulation runtimes and final object pose differences averaged across 5 contact-rich manipulation trajectories involving a planar pushing problem with a T-shaped object and a spherical finger. Column headers show [simulation timestep (ms)]-[integrator]-[\# AOPC points] used in each experiment. The bounding box dimensions for the object are $25\times20 \ \text{cm}^2$, $\Delta d$ denotes the COM distance between end poses of corresponding rollouts from SCM and MJX, while $\Delta \theta$ denotes the difference in 2D rotations. 
}
\label{fig:data_dump}
\vspace{-5mm}
\end{figure*}

We proceed with a characterization of the soft-minimum contact force in \eqref{eq:ssdf_ssdf_force} by comparing it with contact forces from MuJoCo XLA (MJX) \cite{todorov2014convex}. Fig.~\ref{fig:scm_force} visualizes complete contact force profiles from the didactic example with two spheres first introduced in Fig.~\ref{fig:title}, where the small sphere is now allowed to move across the entire x-y plane. It can be seen that MJX's contact force profile involves discontinuities and sharp corners, whereas the SCM force profile is twice differentiable. It can also be seen that with SCM, gradients and Hessians are non-zero even without contact, which is crucial for solving trajectory optimization problems that involve discovering contact \cite{pang2023global}. The idea behind SCM is to make a slight compromise from forward simulation accuracy to instead create contact forces and dynamics that are more suitable for optimization and inference. As can be seen in Fig.~\ref{fig:scm_force}, the two main consequences of this relaxation are:
\begin{itemize}[leftmargin=*, label={\color{ourblue}\textbf{\textbullet}}]
    \item Introduction of an unstable equilibrium state when two spheres are concentric. With sufficiently high contact stiffness, such states are never visited as even the slightest penetration would be sufficient to create large contact forces that push back and resolve the penetration. 
    \item Introduction of contact forces that act at a distance. While these forces may create mild non-physical behavior, they are nevertheless necessary and desirable for optimization as they provide non-zero gradients and Hessians that allow reasoning about the effects of potential contacts from a distance, which facilitates contact discovery \cite{pang2023global}. These imaginary forces can be made arbitrarily small by decreasing the smoothing parameters $\epsilon_{1:3}$ towards zero.
\end{itemize}

\subsection{Characterizations of Simulation Accuracy and Speed}
This section presents experiments that characterize the accuracy and speed of forward simulations under SCM, using MJX as a baseline. Fig.~\ref{fig:fig_sim} shows simulation results for a planar pushing example involving a freely moving T-shaped geometry and a position controlled spherical finger. Five separate control sequences; each tasked with pushing the T-shaped object to a different goal position; were optimized in MJX using the VP-STO algorithm \cite{jankowski2023vp}. These sequences are then rolled out in an open-loop manner in a simulation where MJX's contact model \cite{todorov2014convex} is replaced with SCM instead. Table.~\ref{fig:data_dump} quantifies simulation runtimes and the final object pose differences averaged across all 5 rollouts. Please see the supplementary material for videos of all rollouts. We note that this experiment is designed such that every manipulation trajectory requires establishing and breaking contact at least two times, and combining this with the open-loop nature of the rollouts means the slightest differences in whether a contact sticks or slips can result in large differences at the final object pose. Looking at the table and the associated figure, we can make the following observations:
\begin{itemize} [leftmargin=*, label={\color{ourblue}\textbf{\textbullet}}]
    \item Using SCM is $\sim$ 4 times faster than using MJX's contact model under Euler integration, and has comparable speed under RK4 integration.
    \item As all computations are analytical and vectorized, simulation speed under SCM is independent from the number of AOPC points (i.e., as long as memory limit is not hit) or the number of active contacts.
    \item For reasonable simulation step sizes such as 2ms (i.e., the default simulation stepsize for MJX), the qualitative behavior such as whether contacts stick or slip are similar across both contact models. Under RK4 integration, SCM and MJX trajectories start deviating for simulation step sizes larger than $\sim$ 16ms, and for Euler integration this deviation happens even earlier.  
\end{itemize}
These results highlight that while SCM does make a compromise from forward simulation accuracy, it nevertheless achieves physically plausible behavior sufficient to serve as a description of \say{physics for inference}.

\section{Limitations and Future Work}
The presented approach has two main limitations. The first one is that point-plane contact forces $\bm{\lambda}(\bm{p}, \bm{v}, \bm{p}_i, \bm{n}_i)$ are derived from a spring-damper system \cite{kurtz2023inverse} rather than a convex optimization problem with cone constraints as is the current standard \cite{todorov2012mujoco, drake, castro_sim}. There are two main shortcomings to spring-damper based contact models: i) they require careful tuning of contact hyperparameters (e.g., stiffness, dissipation, stiction) for every new scene to avoid instabilities and unrealistic penetrations, and ii) the individual spring-damper systems for different contacts are unaware of each other as well as other external forces on the system. In contrast, constraints of an optimization-based contact model essentially act analogous to a set of smart spring-dampers (that exert \say{forces} on the KKT system via the dual variables) which are all aware of each other and external forces, as well as scaling their stiffness and damping automatically with inertia \cite{todorov2014convex}. The limitations due to using spring-damper models can be addressed in two ways in future work. The first is to use \eqref{eq:differentiable_contact_points} to obtain a discrete set of contact points that can then be used to transcribe the constraints of any existing optimization-based contact model. The downside of this solution is that it automatically inherits all the shortcomings of optimization-based contact models that this paper tries to overcome. A second and better solution is to replace $\bm{\lambda}(\bm{p}, \bm{v}, \bm{p}_i, \bm{n}_i)$ when computing inverse-dynamics in \eqref{eq:softmin_inv_dyn} with an analytical function $\bm{\lambda}(\bm{p}, \bm{v}, \dot{\bm{v}}, \bm{p}_i, \bm{n}_i)$ that is aware of the desired acceleration $\dot{\mathbf{v}}$. The corresponding forward dynamics can also be easily obtained through an unconstrained optimization problem, which solves for $\dot{\mathbf{v}}$ that matches the applied generalized control force in eq.\ref{eq:softmin_inv_dyn}. A compelling option for $\bm{\lambda}(\bm{p}, \bm{v}, \dot{\bm{v}}, \bm{p}_i, \bm{n}_i)$ is the analytical inverse dynamics formulation used in MuJoCo \cite{todorov2014convex} and Drake \cite{castro_sim}, which involves a projection onto friction cone constraints that can be evaluated via simple geometric arguments in 3D. This constitutes our main direction for future work to address the shortcomings of using spring-damper models.

The second main limitation of the presented approach is that it requires considerable memory, and while batched GPU simulation frameworks such as MJX allow simulating thousand of systems in parallel, this is not possible in the current state of the proposed method. This large memory requirement is a consequence of the simulation time being independent from the number of contacts, which in turn is a consequence of considering all possible point-plane pairs between two AOPCs in \eqref{eq:point_ssdf_force}-\eqref{eq:ssdf_ssdf_force}. Indeed, whenever a simulation involves many different contacts between complex, non-convex meshes, the memory requirements of simulators such as MJX would also scale at least as poorly, with the additional drawback of the simulation runtime growing significantly larger due to increasing number of contact constraints in optimization 
(which is not an issue in the proposed contact model since it purely involves vectorized analytical operations). The main distinction between our approach and simulators like MJX is therefore that in the regime where considering a small number of potential contacts between simple collision primitives is sufficient, their memory requirements scale more favorably (e.g., for quadruped locomotion \cite{caluwaerts2023barkour} where the only contacts are typically between four spherical feet and a planar ground). In such simple cases, our method currently still requires a sufficient number of AOPC points to cover the surfaces of geometric primitives in order to prevent simulation artifacts, and has consequently a larger memory footprint. Therefore our approach and the existing batch simulation approaches are suitable for different regimes and requirements. A promising direction for reducing memory footprint in future work is to explore amortizing SSDF evaluations in \eqref{eq:point_ssdf_force}-\eqref{eq:ssdf_ssdf_force} using neural networks \cite{mueller2022instant} or polynomial basis functions \cite{maric2024online}.

\section{Conclusion}
We presented a formulation of collision detection and rigid body dynamics with contact that purely consist of simple analytical operations that are twice-differentiable. 
The benefits of the proposed approach include: i) obtaining gradients and Hessians through collision detection and contact simulation simply using automatic differentiation, ii) not needing convex decompositions of scene geometry, iii) simulation speed being independent from the number of contacts between geometries, and iv) being able to trivially vectorize simulations of rigid body systems with entirely different geometry (though still sharing the same kinematic structure). While these benefits come at the cost of a slight compromise in physical accuracy (e.g. due to contact forces acting at a distance), we have shown through numerical experiments that the resulting dynamics still maintains sufficient fidelity to serve as an \say{intuitive physics engine} to generate intelligent contact-rich behavior. 
Immediate directions for future work include applications of the proposed contact dynamics to trajectory optimization and system identification problems in robotics.

\bibliographystyle{IEEEtranN} 
\bibliography{scm_iros}

\end{document}